\documentclass[journal]{IEEEtran}

\usepackage{todonotes}
\usepackage{epstopdf}
\usepackage{subfigure}
\usepackage{booktabs} 
\usepackage{float}
\usepackage{rotating}
\usepackage{booktabs,multirow}
\usepackage{hhline}
\usepackage{fancyhdr}
\usepackage{amsmath}
\usepackage{gensymb}
\usepackage{xfrac}
\usepackage{array}
\usepackage{authblk}
\usepackage{siunitx}
\usepackage{epsfig} 
\usepackage{graphicx}
\usepackage{amsmath}
\usepackage{relsize}
\usepackage{color,soul}
\usepackage{amsmath,tabu}

\usepackage{cleveref}
	\crefname{equation}{Eq.}{Eq.}
	\crefname{figure}{Fig.}{Fig.}
	\crefname{table}{table}{tables}

\usepackage[square,numbers,sort&compress]{natbib}

\usepackage{anysize}
\marginsize{2.5cm}{2.5cm}{2cm}{2cm}

\begin{document}

\title{Harnessing Elastic Energy to Transfer Reciprocating Actuation into Rotary Motion}

\author[1]{~Gregor~J.~van~den~Doel}
\author[1]{~Just~L.~Herder}
\author[1]{~Davood~Farhadi}

\affil[1]{Department of Precision and Microsystems Engineering, Delft University of Technology, 2628 CD, Delft, The Netherlands.}

\markboth{Preprint}%
{Preprint}

\maketitle

\begin{abstract}
The ability to convert reciprocating, i.e., alternating, actuation into rotary motion using linkages is hindered fundamentally by their poor torque transmission capability around kinematic singularity configurations. Here, we harness the elastic potential energy of a linear spring attached to the coupler link of four-bar mechanisms to manipulate force transmission around the kinematic singularities. We developed a theoretical model to explore the parameter space for proper force transmission in slider-crank and rocker-crank four-bar kinematics. Finally, we verified the proposed model and methodology by building and testing a macro-scale prototype of a slider-crank mechanism. We expect this approach to enable the development of small-scale rotary engines and robotic devices with closed kinematic chains dealing with serial kinematic singularities, such as linkages and parallel manipulators.
\end{abstract}


\section{Introduction}
\IEEEPARstart{A}{ctuators} are a fundamental component for robots and machines to accomplish the tasks for which they are designed. Over the past few decades, there has been unprecedented progress in developing new actuation technologies at different length scales. Most of these actuators, particularly at the small scale, are primarily available in reciprocating motion, e.g., rectilinear and rotational, with limited travel range. However, converting a reciprocating actuation into a full-cycle rotary motion has been a long-standing challenge for many applications and hampered the development of advanced robots and mechanical systems. 

Application examples include (i) micro-robotics where a reciprocating actuation scheme \cite{bishop2015design, zhang2019robotic, yang202088} can be tailored for a rotary motion in micro air vehicles\cite{floreano2015science, dudley2015micro,kumar2012opportunities}, legged locomotion \cite{floyd2008design, saranli2001rhex, carbone2005legged} and wheeled locomotion \cite{wu2018wheeled, campion2008wheeled, ortigoza2012wheeled}, (ii) as a motion converter transmission mechanism in compliant mechanical wristwatches to accommodate the type of motion, i.e., conversion between reciprocating and rotary motions, of time dials and mainspring \cite{wessels2017reciprocating, machekposhti2018compliant, farhadi2019frequency, rubbert2016isotropic, vardi2018theory, machekposhti2018statically}, (iii) driving micro-scale engines using linear actuators for pop-up mirrors, optical switches, and micropositioners \cite{garcia1995surface, wu1997micromachining, gad2001mems, tanner2000mems}, and (iv) impart directionality for nanoscale rotary motors \cite{ketterer2016nanoscale, shi2022sustained, pumm2022dna} that mimic ATP (adenosine triphosphate) synthase \cite{yoshida2001atp, srivastava2018high} or the flagellar motors \cite{minamino2008molecular}.

\begin{figure*}[t!]
	\centering
	\includegraphics[width=0.85\linewidth]{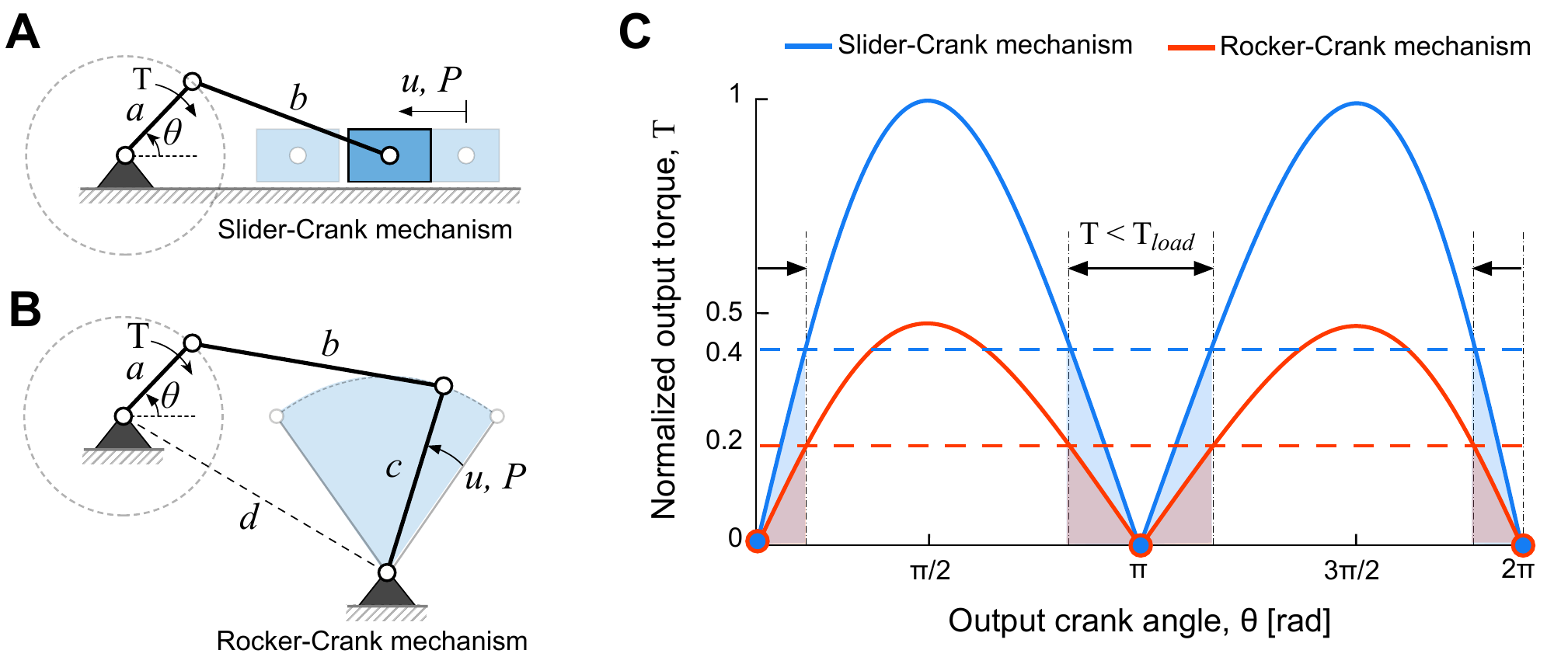}
	\caption{Kinematic singularity issue of four-bar mechanisms in converting reciprocating actuation into a full cycle continuous rotary motion. (A) Slider-Crank mechanism, (B) Rocker-Crank mechanism, (C) Normalized output torque $T$ vs. output crank angle $\theta$, where $b/a=6$ is the dimensionless parameter for coupler link in both mechanisms. In addition, $c/a=2$ and $d/a=6.2$ are the other length ratios in the case of Rocker-Crank mechanism. The dashed lines indicate a minimum output torque required based on an application load, e.g., $40 \%$ of the maximum transmitted torque.}
	\label{Figure1} 
\end{figure*}

An ideal approach to convert a reciprocating actuation into a continuous rotary motion is to use the kinematics of four-bar linkages, e.g., slider-crank and rocker-crank mechanisms shown in Fig. \ref{Figure1}A and B. However, these mechanisms can not be utilized since they suffer from the first kind of kinematic singularity, also known as the serial kinematic singularity. This configuration corresponds to those in which the coefficient of output crank velocity $d\theta/dt$ becomes zero in the velocity equation \cite{gosselin1990singularity, kieffer1994differential, gosselin2015singularity, garcia2018feasibility}. This situation induces a zero velocity for a reciprocating input, i.e., $du/dt=0$, regardless of the value of $d\theta/dt$. As a result, the reciprocating input force, or torque in the case of rocker-crank kinematics, will not generate any output torque at the crank link. In other words, the mechanism can not be controlled with one input due to instantaneous changes in its degrees of freedom (DoF) at singularity configuration. Therefore, it is kinematically impossible to convert a reciprocating actuation into a continuous rotary motion through linkage-based mechanisms.

There are two main approaches to obviate the singularity issues in kinematic design in the prior art. A centuries-old approach is adding a flywheel at the crank link. When there is sufficient speed, the flywheel stores kinetic energy and exploits that at singularity configuration where input energy is insufficient. However, this solution imposes several limitations. For instance, the additional weight of the flywheel is, in many cases, undesirable for macro-scale devices. In addition, the solution can not be scaled down for application in micromachines \cite{garcia1995surface} and nanorobotics \cite{marras2015programmable}, since the kinetic energy stored in the small systems decreases very rapidly with size, i.e., with power five of scale size. Furthermore, there are other forces at the small scale that are more dominant than kinetic energy, e.g., friction, adhesion, and Van der Waals forces \cite{kendall1994adhesion}.

The second approach to eliminate the singularity issue in motion conversion is using an additional actuator with a phase shift from a primary actuator \cite{garcia1995surface}. However, this solution requires a control unit, additional infrastructure, and additional space, resulting in system and manufacturing complexity.

In this work, we propose a new approach that harnesses the elastic potential energy of a single spring to obviate kinematic singularity issues in four-bar linkages, e.g., slider-crank and rocker-crank mechanisms. We first show that, by adding a spring to the coupler link of the four-bar kinematics, the system can store elastic deformation and release stored elastic potential energy at singular configurations. We study via a combination of experiments and theoretical analysis the force transmission capability of the system. Then we demonstrate the possible solutions in both slider-crank and rocker-crank mechanisms. The proposed solution provides opportunities for simplification (e.g., reducing the number of actuators and inertia) and scalability of robotic systems that suffer from kinematic singularity issues.

\section{Methodology and Results}
\subsection{Force Transmission Near Singularity}
The Force transmission capability of slider-crank and rocker-crank linkages are normalized and depicted in Fig. \ref{Figure1}C. For both mechanisms, input load $P$ (i.e., an input force or torque) is constant. The transmitted torque $T$ at the output crank link can be calculated with the corresponding input displacements $u$ and the angle of crank link $\theta$, as follows

\begin{equation}
    T (\theta) = P \frac{du}{d\theta}.
\end{equation}

As can be seen, poor force transmission capability is not only limited to the singularity configurations, $\theta$ of $0^{\circ}$, and $\pi$, but also the regions near these configurations. The problem becomes evident when a minimum output torque is required based on an application load $T_{load}$, e.g., a required torque that is at least $40 \%$ of a maximum transmitted torque, shown as black dashed lines in Fig. \ref{Figure1}C. As a result, these mechanisms cannot transmit the required torque at the output, i.e., $T<T_{load}$, due to their kinematics. Problematic regions are smaller if the required output torque is low. Therefore, a motion transmission problem around singularity can be defined as \textit{the region where the transmitted torque from an input is smaller than the required output torque, caused by the kinematics in combination with the required output load}.

\subsection{Harnessing Elastic Deformation to Obviate the Singularity Problem}
\begin{figure}[ht!]
	\centering
	\includegraphics[width=0.85\linewidth]{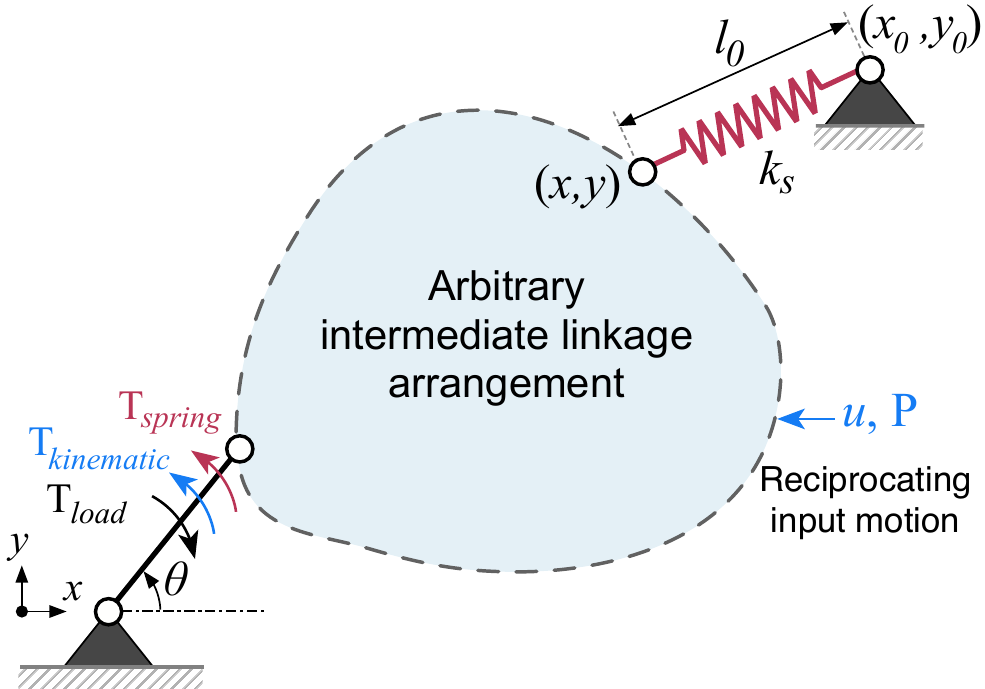}
	\caption{Schematic of the proposed methodology. The mechanism harnesses elastic deformation to obviate force transmission problems around serial kinematic singularity configurations.}
	\label{Figure2}
\end{figure}

To obviate the singularity problem, we consider a single spring attached between the ground and a point on a motion converter mechanism, with stiffness $K_s$ and relaxed length $l_0$ (Fig. \ref{Figure2}). The spring attachment point must be chosen such that the reciprocating input actuation determines spring deformation, and the elastic potential energy of the spring can be tailored to provide an alternative passive torque on a crank link at singularity configurations.

\begin{figure*}[t!]
	\centering
	\includegraphics[width=0.95\linewidth]{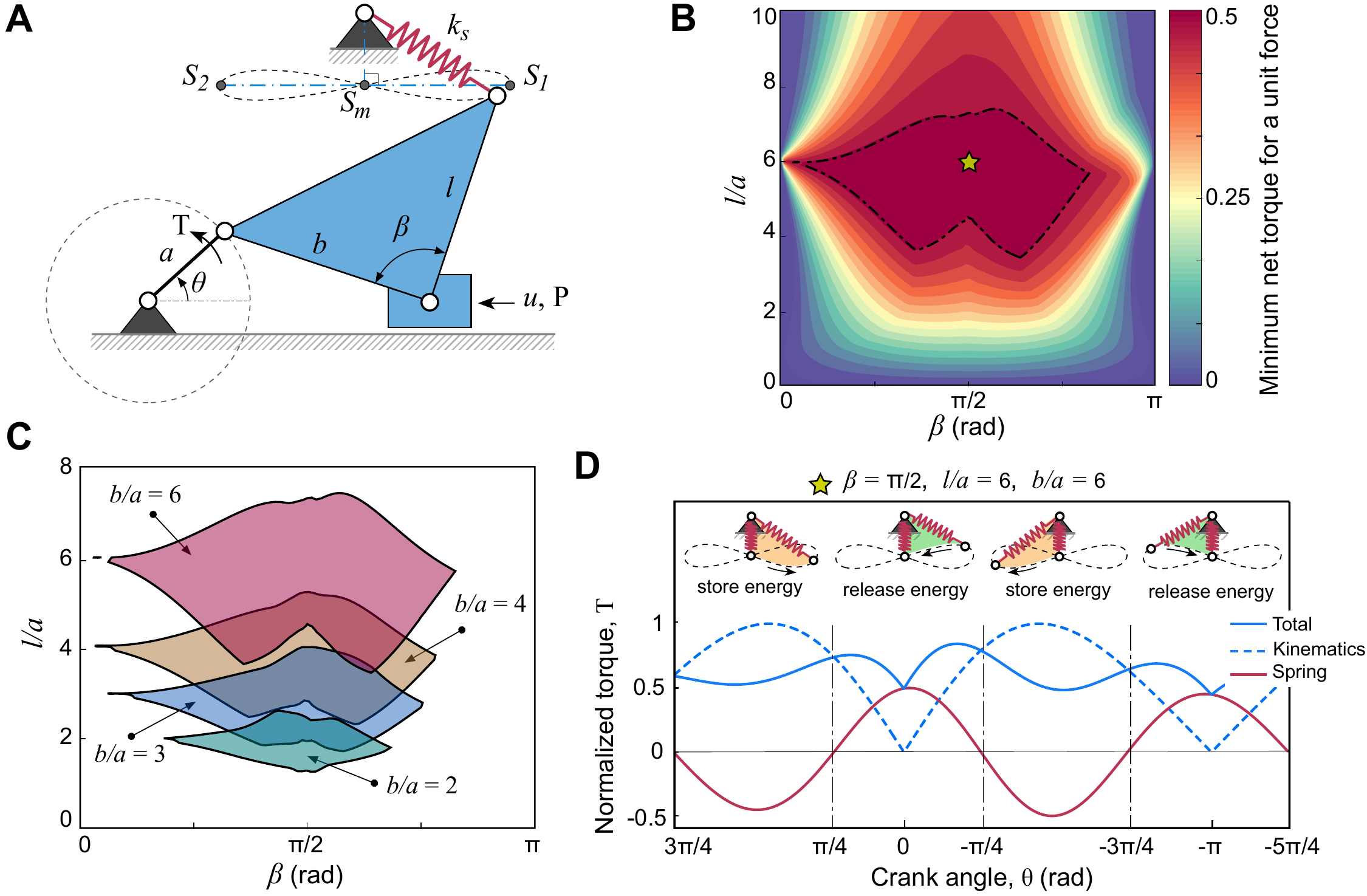}
	\caption{The proposed method and theoretical model for harnessing elastic deformation to obviate kinematic singularity issues in slider-crank mechanisms. (A) Schematic representation of the mechanism with additional geometric parameters, $\beta$ and $l$, for the coupler link attached with a translational spring $k_s$. (B) Theoretically predicted evaluation of the minimum net torque at the crank as a function of geometrical parameters $l/a$ and $\beta$, considering a unit reciprocating input force $P$ at the slider link. The dashed line indicates the optimum design area, where the mechanism achieves a minimum net torque that is $40\%$ of the maximum transmitted torque. (C) The optimum design area for different $b/a$ values. (D) Normalized output torque $T$ at the crank link is evaluated as a function of its angle $\theta$. Total torque is depicted with a solid blue line, and contributions from kinematics and spring are indicated with dashed blue and solid red lines, respectively. In this case, $b/a=6$, $\beta =\pi/2$ and $l/a=6$.}
	\label{Figure3}
\end{figure*}

The spring characteristics, i.e., relaxed length $l_0$ and stiffness $K_s$, can be determined based on the required output load and the extension cycle of the spring. Spring constant $K_s$ can be calculated from the difference between the maximum $l_{max}$ and minimum $l_{min}$ extended length of the spring and can be written as
\begin{equation}
    K_{s}= \frac{2 W_s}{(l_{max}-l_{min})^2},
    \label{springconstant}
\end{equation}
where, $W_s=T_{load}\theta_s$ is the amount of energy required to move an output load over the largest unfavorable region $\theta_s$, i.e., regions where $T<T_{load}$. This calculation ensures enough energy can be stored in the spring to move the mechanism through the singularity configurations. In addition, the relaxed spring length is equal to the minimal distance between the connection points to eliminate pretension in the system.

By defining the Lagrangian of the mechanism, including the spring, the total torque acting on the crank link can be given by
\begin{equation}
    T (\theta) = P \frac{du}{d\theta}-\frac{1}{2}K_s\frac{d}{d\theta}(l-l_0)^2,
    \label{Lag}
\end{equation}
where the first term corresponds to the torque generated by the kinematics of an arbitrary intermediate linkage arrangement, and the second term corresponds to the torque generated by the spring. In this equation, length $l$ corresponds to the deflected length of the spring and can be given by
\begin{equation}
    l = \sqrt{(x-x_0)^2+(y-y_0)^2},
\end{equation}
where $(x_0,y_0)$, and $(x,y)$ correspond to the coordinates of the grounding and attachment points of the spring, respectively. In Eq. \ref{Lag}, the term for kinetic energy is excluded in the Lagrangian of the mechanism since the focus of this work was on the quasi-static situations, where the impact of singularity is dominant.

The proposed method is implemented on the slider-crank and rocker-crank four-bar mechanisms. For a full cycle rotation, these mechanisms face two singularity configurations. Therefore, the spring placement is critical for the feasibility of the method. The attached spring should release the energy two times during a complete motion cycle to obviate the singularity problem and result in a continuous rotary motion at a crank link. To achieve this, two conditions must be satisfied: (1) The elastic potential energy cycle of the spring must have two transition points, i.e., points where a spring changes from storing energy state into releasing energy state, one for each singularity configuration. (2) The singularity configurations must be in the energy release parts of the elastic potential energy cycle. Ideally, the spring must be connected between the available links (e.g., crank, coupler, rocker, or slider) and the ground link to avoid using additional links or linkages.  

\begin{figure*}[ht!]
	\centering
	\includegraphics[width=0.95\linewidth]{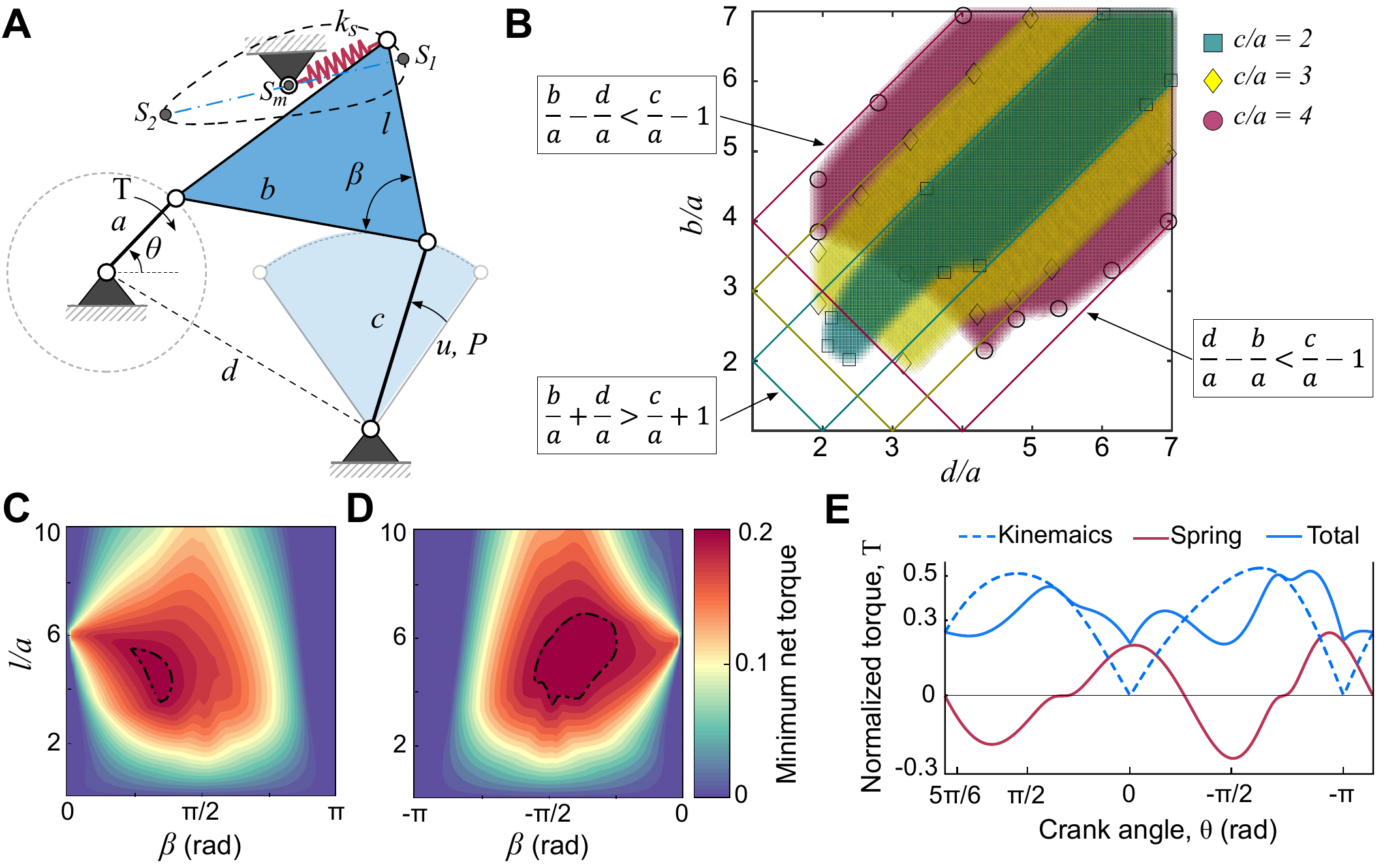}
	\caption{The results for rocker-crank mechanisms. (A) Schematic representation of the mechanism with additional geometric parameters, $\beta$ and $l$, for the coupler link attached with a translational spring $k_s$. (B) Solution space for rocker-crank mechanisms with different transmission ratios, i.e., $c/a= 2, 3, 4$. Colored areas represent cases with a minimum net torque $40\%$ of the maximum transmitted torque. The solid lines represent areas that satisfy the Grashof criteria for a continuous crank rotation. (C) Theoretically predicted evaluation of the minimum net torque at the crank as a function of geometrical parameters $l/a$ and $\beta$, considering a unit reciprocating input $P$ and parameters resulting in a clockwise torque at the crank. (D) Results for a counterclockwise torque at the crank. For C and D, $b/a=6$, $d/a=6.2$, $c/a=2$, and the dashed lines indicate areas where the linkage achieves a minimum net torque that is $40\%$ of the maximum transmitted torque. (E) Normalized output torque $T$ at the crank link is evaluated as a function of its angle $\theta$. Total torque is depicted with a solid blue line, and contributions from kinematics and spring are indicated with dashed blue and solid red lines, respectively. In this case, $\beta =\pi/3$, $l/a=4.4$, and mechanism dimensions are the same as in C and D.}
	\label{Figure4}
\end{figure*}

Every point on a four-bar mechanism travels through a path during a motion conversion, where the path's shape determines the possibility of providing two transition points for a translational spring.
Paths on a crank link are circular and provide only one transition point for spring throughout a complete cycle, not satisfying the first condition. Points on the slider and the rocker links provide two transition points, fulfilling the first condition. However, these transition points are at the same position as singularity configurations, where it stores energy in an unfavorable region. Therefore, attaching a spring to the slider or rocker links does not satisfy the second condition.

In contrast, a coupler link has various high-degree paths, e.g., pseudo ellipse, curves with a crunode, and cusp. These curves could potentially provide two transition points with energy-releasing parts of the spring away from singularity configurations. In the following, we will discuss the results for both the slider-crank and rocker-crank kinematics.

\subsubsection{Slider-Crank mechanism} Results for a slider-crank mechanism are shown in Fig. \ref{Figure3}. The attachment point of the spring is generalized by defining two additional geometric parameters, length $l$ and angle $\beta$, on an extended coupler link (Fig. \ref{Figure3}A). Points with the longest relative distance on a coupler curve define the transition points, e.g., points $S_1$ and $S_2$. The midpoint of the line between the two transition points determines the grounding location of the spring, demonstrated as point $S_m$. If the path of the coupler link intersects with the midpoint, the perpendicular bisector of the transition points determines the grounding point (Fig. \ref{Figure3}A). In addition, to prevent undesired interference between the spring and mechanism, the minimal distance between the two connection points of the spring must not be smaller than the unstretched length of spring $l_0$. 

\begin{figure*}[t!]
	\centering
	\includegraphics[width=1\linewidth]{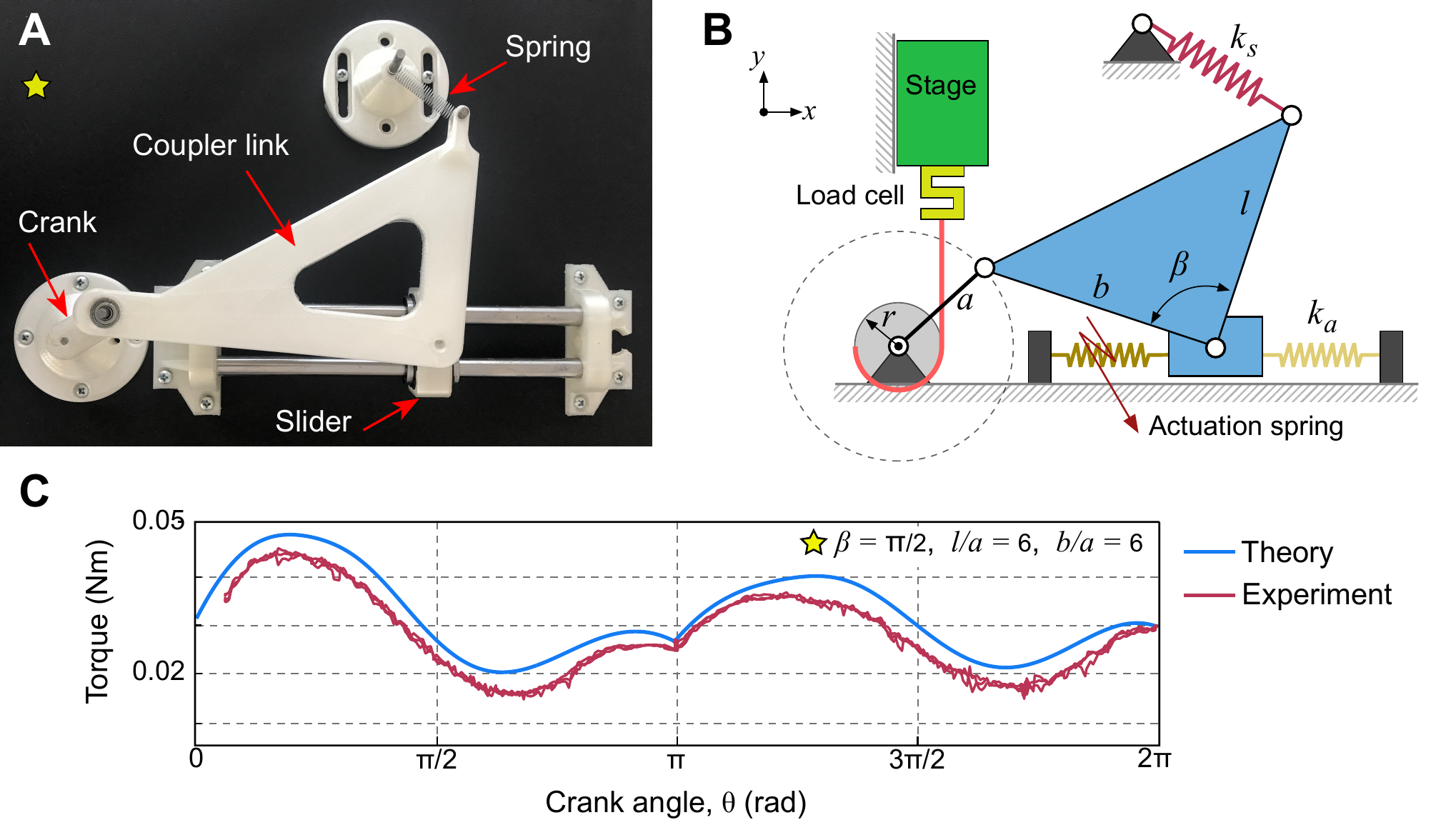}
	\caption{Performance characterization of the design example. (A) Photo of the prototype indicated with different parts. (B) Schematic of the measurement setup. (C) Results from both measurement and the theoretical model.}
	\label{Figure5}
\end{figure*}

In Fig. \ref{Figure3}B, we report the evaluation of minimum net torque at the output crank, for a clockwise rotation, as a function of geometric parameters $l/a$ and $\beta$ for a slider-crank mechanism with links ratio of $b/a=6$. The results indicate that, while the minimum net torque is $40\%$ of maximum transmitted torque, there is a wide range of solutions for the optimum geometric parameters, making the design insensitive to tolerances. The connection points between $\pi<\beta<2\pi$ behave similarly and result in a negative torque value, i.e., generating a counterclockwise crank rotation. Therefore, the mechanism and spring arrangement must be mirrored with respect to the horizontal ground link if a counterclockwise torque is required.

The optimum geometric parameter $l/a$ scales similarly with links ratio $b/a$ (Fig. \ref{Figure3}C). Differently, the solution range for geometric parameter $\beta$ reduces exponentially when decreasing links ratio $b/a$. In addition, there is theoretically no solution for links ratio $b/a\leq1$ since this ratio does not satisfy the Grashof criterion for a continuous crank rotation \cite{barker1985complete}. 

Guided by these results, we choose $b/a=6$, $l/a=6$, $\beta=\pi/2$ and evaluate the normalized output torque, including contributions from kinematics and spring energy, as a function of crank angle (Fig. \ref{Figure3}D). The spring stores energy in two alternating quarters and releases it in the next following quarters, where the spring energy forces the mechanism to snap through each singularity configuration. 

\subsubsection{Rocker-Crank mechanism} In Fig. \ref{Figure4}, we show results for the rocker-crank mechanisms. Similar to the slider-crank mechanisms, the attachment point of the spring is generalized by defining two additional geometric parameters, length $l$ and angle $\beta$, on an extended coupler link (Fig. \ref{Figure4}A). The midpoint $S_m$ does not intersect with the coupler curve, and thus it represents the grounding location for the spring. 

In Fig. \ref{Figure4}B, we demonstrated the solution space of rocker-crank mechanisms. We divide the lengths of the links $b$, $c$, and $d$ by the length of the crank link $a$ to define three non-dimensional parameters. Therefore, every possible combination can be represented by a particular set of these non-dimensional parameters, $b/a$, $c/a$, and $d/a$. Here, the non-dimensional parameter $c/a$ is the transmission ratio between the rocker and the crank links. Grashof theorem states two conditions for a rocker-crank mechanism, i.e., a four-bar mechanism with at least one revolving link. These are (1) the crank (i.e., link with length $a$) being the shortest link in the mechanism, and (2) the sum of the shortest and the longest links is less than the sum of the length of the remaining two links \cite{barker1985complete}. For a given transmission ratio $c/a$, these conditions give three inequalities, each corresponding to one particular link being the longest link in the mechanism, except the crank link. The area illustrated by these inequalities shows the solution space for the rocker-crank mechanism. We then use Eq. \ref{Lag} to evaluate every combination in solution space and outlined feasible dimensions that would result in a minimum net torque $40\%$ of maximum transmitted torque. Area outlined in colored markers indicate feasible solutions for each transmission ratio. It is evident that not all rocker-crank mechanisms meet the torque requirement at the crank link.

The contour plots depicted in Fig. \ref{Figure4}C and D show the effect of the geometric parameters $l/a$ and $\beta$ on the minimum net torque behaviour at the output crank for both clockwise (Fig. \ref{Figure4}B) and counterclockwise (Fig. \ref{Figure4}C) directions. In contrast to the slider-crank mechanism, the optimum solution area in the rocker-crank mechanism differs depending on crank direction, which is caused by the asymmetrical trajectory of the coupler curve. 

In Fig. \ref{Figure4}E, we evaluate the normalized output torque, including contribution from kinematics and spring energy, as a function of crank angle for a set of design parameters derived from parametric studies. 

\subsection{Experimental Results}
Guided by these results, we choose to build a slider-crank mechanism with crank link $a=30\:mm$, coupler link $b=180 \:mm$, $\beta = \pi/2$, and extended link $l=126 \:mm$; and a spring with $k_s=0.0568 \: N/mm$ and $l_0=20 \:mm$ (Fig. \ref{Figure5}A, and Movie S1). We made a test setup to evaluate the output torque performance experimentally, schematic shown in Fig. \ref{Figure5}B. We fix the mechanism on a flat surface in the $x-y$ plane (note that gravity acts in the $z$ direction and does not influence the measurement) and connect the crank link to a load cell (FUTEK LSB200). A nylon rope, wound around crank pin with a radius of  $r=7 \:mm$, connects the load cell to the crank link. The load cell is mounted on a rectilinear stage, which moves in $y$ direction and allows for clockwise crank rotation. The travel distance of the stage is then converted into an angular displacement of the crank. 

In the experiment, we chose to actuate the slider directly with two linear springs, each with  $k_a=0.0188 \: N/mm $, and activated one at a time. This is done manually by connecting one end of the actuation spring with a rod and disconnecting the other spring. Using the linear stage, we positioned the crank close to $0$ degrees position and applied a pre-load with the left actuation springs. At this configuration, the actuation spring on the right side is disconnected. We then start the measurement by programming the linear stage to move the crank towards $180$ degrees in the clockwise direction and measure the reaction force along the way. In the second step, we first stopped the linear stage, disconnected the left actuation spring, and pre-loaded the actuation spring on the right side. We then continue the force measurement by rotating the crank link, using the linear stage, from $180$ degree towards $0$ degree angle clockwise.

The maximum pretension applied to the actuation springs is approximately $47 \:mm$. The torque-angular displacement results from both the experiment and theoretical model are shown in Fig. \ref{Figure5}C. The theoretical model is updated with spring force as an input to compare the results. As shown in Fig. \ref{Figure5}C, there is a good agreement between the two sets of data, confirming the validity of the theoretical model and the proposed method. It is important to note that friction, e.g., at the slider and the rotary hinges, is not considered in the theoretical model. This explains the downward shift of the measured torque at the crank link.

\section{Discussion and outlook}
The demonstration of harnessing elastic potential energy in four-bar linkages makes a promising route to mechanically control mechanisms and robotic systems \cite{daniali1995singularity} at their serial kinematic singularity. In addition, integration of a spring into transmission mechanism for reciprocating actuator concepts means that rotary engine could be created without control complexity at the small scale, e.g., DNA-based rotary apparatus that mimic the biological rotary motors such as bacterial flagellar motor \cite{ketterer2016nanoscale, yonekura2003complete, huang2021integrated}. 

Although we use the coil spring as the elastic element, the design could be simplified by replacing the spring with a compliant alternative. This could be implemented by grounding the coupler link with a compliant joint \cite{farhadi2015review}.

The implementation of the methodology developed here is not without limitations. The method does not include mechanisms that have multiple overlapping serial kinematic singularities during a full cyclic motion, where each corresponds to different motion direction. An example of this scenario can be shown in double-slider mechanisms, i.e., combination of four links having two turning pairs and two sliding pairs. Considering the motion directions of the two sliders to be perpendicular to each other, one moving in a horizontal direction while the other moving in a vertical direction, there will be two overlapping serial kinematic singularities. Both will occur when the horizontal slider crosses the vertical slider's motion axis; once when it moves from right to left, and once when it moves from left to right. It is not possible to employ the proposed method as the coupler curves are limited to elliptical shapes without crunode and cusp points.


\section{Conclusion}
In summary, we have shown that elastic potential energy can be harnessed to overcome serial kinematic singularity problems in four-bar linkages with reciprocating, i.e., alternating, input and rotary crank output.

The proposed solution relies on a passive principle using a single spring connected between the coupler link in the four-bar linkages and the base, eliminating the need for additional actuators and control circuitry. More specifically, we have demonstrated this by creating an elastic potential energy cycle with two transition points and locating the energy release around the serial kinematic singularity configurations.

Although in this work, we have presented the prototype at macro scale, it is important to point out that our approach can be extended to design motion transmission mechanism over a wide range of length scales as long as the energy stored in the spring is enough to overcome both the dissipated energy and the required output load.

\bibliography{TRSingularity}{}

\end{document}